\documentclass[times,twocolumn,final,authoryear]{article}
\usepackage{simpleConference}
\usepackage{times}
\usepackage{graphicx}
\usepackage{amssymb}
\usepackage{url,hyperref}
\usepackage{xcolor}

\usepackage{natbib}
\usepackage[ngerman, english]{babel}
\usepackage{graphicx}
\usepackage{multirow}
\usepackage{array}
\usepackage{enumitem}
\usepackage{amsmath} 
\setitemize{noitemsep,topsep=0.5pt,parsep=0.5pt,partopsep=0.5pt}
\newcolumntype{P}[1]{>{\centering\arraybackslash}p{#1}}
\usepackage{booktabs}

\usepackage{authblk}
\author[1]{Chengwei Wei}
\author[1]{Runqi Pang}
\author[1]{C.-C. Jay Kuo}
\affil[1]{University of Southern California, Los Angeles, California, USA}

{
    \makeatletter
    \renewcommand\AB@affilsepx{: \protect\Affilfont}
    \makeatother

    \makeatletter
    \renewcommand\AB@affilsepx{, \protect\Affilfont}
    \makeatother

    \affil[ ]{\texttt{chengwei@usc.edu}}
    
}

\begin{document}

\title{GWPT: A Green Word-Embedding-based POS Tagger}


\maketitle
\thispagestyle{empty}

\begin{abstract}
As a fundamental tool for natural language processing (NLP), the
part-of-speech (POS) tagger assigns the POS label to each word in a
sentence. A novel lightweight POS tagger based on word embeddings is
proposed and named GWPT (green word-embedding-based POS tagger) in this
work. Following the green learning (GL) methodology, GWPT contains
three modules in cascade: 1) representation learning, 2) feature
learning, and 3) decision learning modules. The main novelty of GWPT
lies in representation learning. It uses non-contextual or contextual
word embeddings, partitions embedding dimension indices into low-,
medium-, and high-frequency sets, and represents them with different
N-grams. It is shown by experimental results that GWPT offers
state-of-the-art accuracies with fewer model parameters and
significantly lower computational complexity in both training and
inference as compared with deep-learning-based methods.
\end{abstract}


\section{Introduction}\label{sec:intro}

Part of speech (POS) tagging is one of the classical sequence labeling
tasks. It aims to tag every word of a sentence with its POS attribute.
As POS offers a syntactic attribute of words, POS tagging is useful for
many downstream tasks, such as speech recognition, syntactic parsing,
and machine translation.  POS tagging has been successfully solved with
complex sequence-to-sequence models based on deep-learning (DL)
technology, such as LSTM \citep{wang2015part, wang2015unified,
bohnet2018morphosyntactic} and Transformers \citep{li2021part}. 

Recently, the NLP landscape is dominated by Large Language Models (LLMs)
\citep{wei2023overview, brown2020language, openai2023gpt}.  There is a
perception that the rise of LLMs, which excel in many applications, has
shifted the focus. However, LLMs are based on generative pre-trained
transformers (GPTs).  They are challenged by hallucination, reliability,
huge computational complexity, and lack of incremental learning
capabilities.  To tackle these deficiencies, an alternative approach
could be the decomposition of a generic LLM to several smaller
domain-specific mid-size language models (MLM) that have a "multi-modal
interface" to handle textual or visual input/output and a "knowledge
core" implemented by knowledge graphs (KGs) for knowledge
representation, data mining, and incremental learning. Such a modular
design could improve the interpretability, reliability, computational
complexity, and incremental capability of next-generation MLMs with
justifiable and logical reasoning. In this direction, POS tagging is
still a valuable step in building interpretable NLP models. Furthermore,
there is a need for lightweight high-performance POS taggers to offer
efficiency while ensuring efficacy for downstream tasks.  

In this work, we propose a novel word-embedding-based POS tagger, called
GWPT, to meet this demand.  Following the green learning (GL)
methodology \citep{kuo2022green}, GWPT contains three cascaded modules:
1) representation learning, 2) feature learning, and 3) decision
learning. The last two modules of GWPT adopt the standard procedures,
i.e., the discriminant feature test (DFT) \citep{yang2022supervised} for
feature selection and the XGBoost classifier in making POS prediction.
The main novelty of GWPT lies in representation learning.  

GWPT derives the representation of a word based on its embedding. Both
non-contextual embeddings (e.g., fastText) and contextual embeddings
(e.g., BERT) can be used. GWPT partitions dimension indices into low-,
mid-, and high-frequency three sets. It discards dimension indices in
the low-frequency set and considers the N-gram representation for
dimension indices in the mid- and high-frequency sets. Furthermore, the
final word features are selected from a subset of word representations
using supervised learning. It helps mitigate the adverse impacts of
noise or irrelevant features for POS tagging tasks and reduce
computational costs simultaneously.  Extensive experiments are conducted
for performance benchmarking between GWPT and several DL-based POS
taggers.  As compared with DL-based POS taggers, GWPT offers highly
competitive tagging accuracy with fewer model parameters and
significantly lower complexity in training and inference. 

There are two main contributions of this work.
\begin{itemize}
\item A new efficient representation method for POS tagging derived from
word embeddings is proposed.  It discards low-frequency dimension
indices and adopts N-gram representations for those in the mid- and
high-frequency sets to enhance the overall effectiveness of the proposed
GWPT method. 
\item Extensive POS tagging experiments are conducted to evaluate
tagging accuracy, model sizes, and computational complexity of several
benchmarking methods. GWPT offers competitive tagging accuracy with
smaller model sizes and significantly reduced complexity. 
\end{itemize}

The rest of this paper is organized as follows. Related previous work is
reviewed in Sec. \ref{sec:RelatedWork}. The GWPT method is described in
Sec. \ref{sec:Method}. Experimental results are presented in Sec.
\ref{sec:experiments}.  Concluding remarks are given in Sec.
\ref{sec:conclu}. 

\section{Related Work}\label{sec:RelatedWork}

POS tagging methods can be categorized into rule-based,
statistical-based, and DL-based three approaches as elaborated below. 

{\bf Rule-based approach.} Rule-based POS tagging methods
\citep{brill1992simple, eric1994some, chiche2022part} utilize
pre-defined linguistic rules to assign POS tags to words in sentences.
Generally, a rule-based POS tagger initially assigns each word its most
likely POS using a dictionary derived from a large tagged corpus without
considering its context. Then, it applies rules to narrow down and
determine the final POS for each word. These rules are created by
linguistic experts or corpus based on linguistic features of a language,
such as lexical, morphological, and syntactical patterns.  For example,
switching the POS tag from VBN to VBD when the preceding word is
capitalized \citep{brill1992simple}. While rule-based methods offer
simplicity and interpretability, their performance is inadequate in the face
of complex and ambiguous instances of a language. 

{\bf Statistical-based approach.} Statistical-based POS tagging methods,
also called stochastic tagging, utilize annotated training corpora to
learn the statistical relationship between words and their associated
POS tags. Specifically, they disambiguate words by considering the
probability of a word occurring with a specific tag in a given context.
Statistical-based POS tagging often adopts the hidden Markov model (HMM)
\citep{kim1999hmm, lee2000lexicalized, van2009infinite}, where POS tags
are the hidden states and words in a sentence sequence serve as
observations. HMM-based POS taggers aim to learn the transition
probability (i.e., the probability of one POS tag succeeding another)
and the emission probability (i.e., the probability of a word being
emitted from a specific POS tag) from annotated training corpora.
Besides HMM, other statistical models have also been considered such as
the maximum entropy model \citep{ratnaparkhi1996maximum,
zhao2002chinese} and the conditional random fields (CRF)
\citep{agarwal2006part, pvs2007part, silfverberg2014part}. 

{\bf DL-based approach.} DL-based POS tagging methods have gained
popularity as their ability to capture linguistic patterns from a large
number of training data and achieve high performance.  Common models
include recurrent neural networks (RNN) \citep{wang2015part,
wang2015unified, bohnet2018morphosyntactic} and transformers
\citep{li2021part}. The performance of DL-based taggers can be enhanced
by integrating with other techniques such as character embeddings,
adversarial training, or rule-based pre-processing.  DL-based POS
taggers outperform rule-based and statistical-based methods at
substantially higher computational and storage costs.  Recently, large
language models (LLMs) \citep{brown2020language, openai2023gpt} can
manage POS tagging implicitly and address downstream NLP tasks directly. 

Our work follows the GL paradigm \citep{kuo2022green,
kuo2016understanding, kuo2019interpretable}.  GL aims to address the
high computational and storage costs of the DL paradigm while providing
competitive performance at the same time. GL has neither neurons nor
networks.  It is characterized by low carbon footprints, lightweight
models, low computational complexity, and logical transparency. It offers
energy-efficient solutions in AI chips, mobile/edge devices, and data
centers. GL methods have been successfully developed for quite a few
image processing and computer vision tasks \citep{kuo2018data,
chen2020pixelhop++, chen2021defakehop, zhang2020pointhop, kadam2022r}.
In this work, we demonstrate the design of a green POS tagger, GWPT, and
conduct performance benchmarking between GWPT and DL-based methods. 

\begin{figure*}[htb]
\centerline{\includegraphics[width=0.78\linewidth, height=7.5cm]{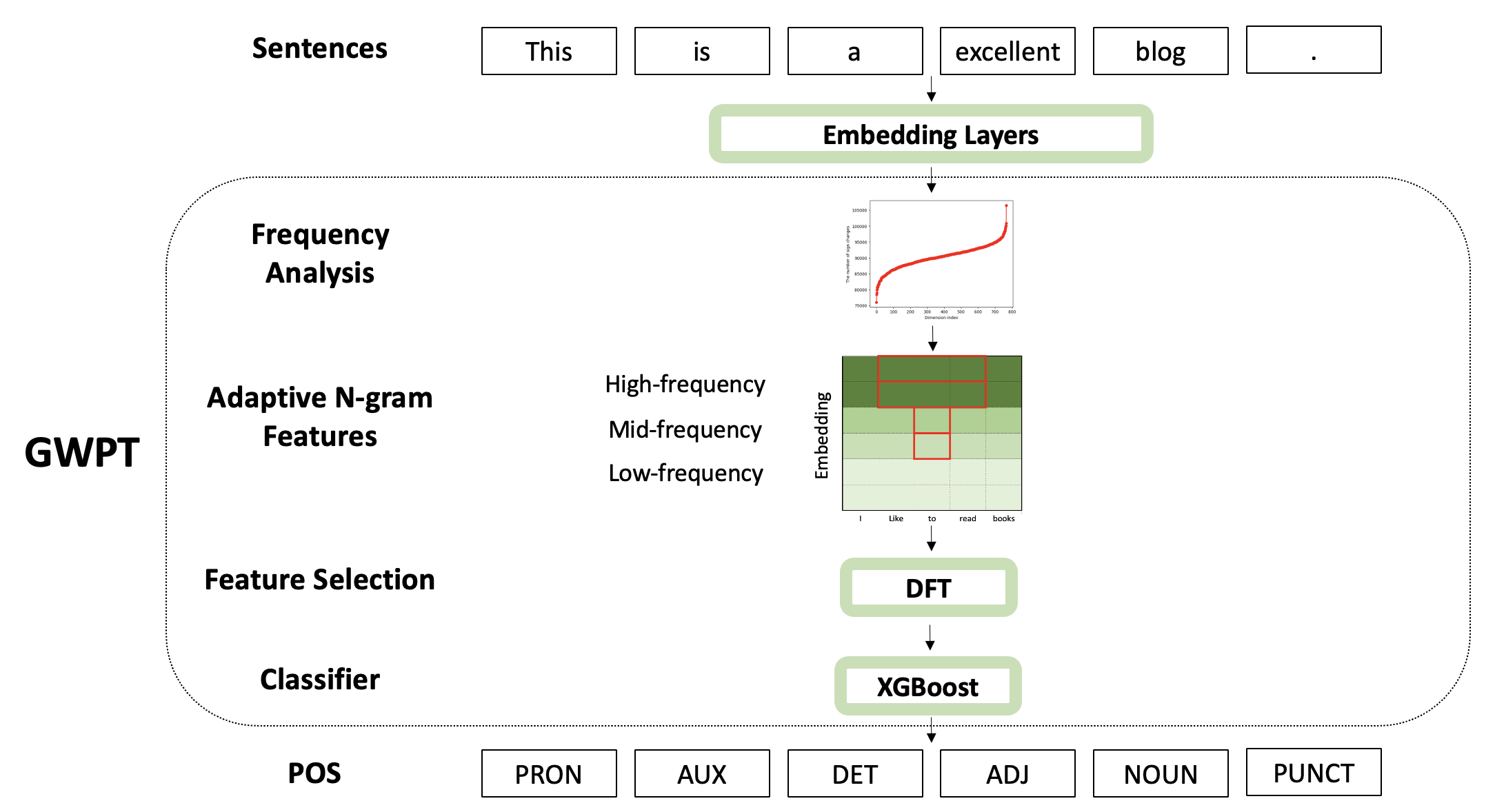}}
\caption{The system diagram of the GWPT method.}\label{fig:pipeline}
\end{figure*}

\section{Proposed GWPT Method}\label{sec:Method}

The system diagram of GWPT is depicted in Fig. \ref{fig:pipeline}. It
contains four steps.  Steps 1 and 2 belong to the unsupervised
representation learning module. Steps 3 and 4 correspond to the
supervised feature learning and the supervised decision learning
modules, respectively. 
\begin{enumerate}
\item {\em Frequency Analysis of Embedding Dimensions.} We analyze the
frequency of each word embedding dimension and partition word embedding
dimension indices into low-, mid-, and high-frequency sets. 
\item {\em Concise Representation with Adaptive N-grams.} We adopt
adaptive N-grams to each word embedding dimension based on their
frequency analysis.  The red block in Fig. \ref{fig:pipeline} shows the
N-gram ranges associated with word embedding dimensions of different
frequencies.  The adaptive N-gram design captures the essential
contextual information for accurate POS prediction. 
\item {\em Discriminant Feature Selection.} The dimension of
concatenated N-grams of a word is still large. We adopt a
semi-supervised feature extraction tool, DFT \citep{yang2022supervised},
to select features of higher discriminant power. 
\item {\em Classification for POS Labels.} We perform the word-based POS classification
task using an XGBoost classifier. 
\end{enumerate}
These four steps are elaborated below.

\subsection{Frequency Analysis of Embedding Dimensions}

Consider an $L$-dimension word embedding scheme, which can be
contextual- or non-contextual-based. We denote each dimension with
$D_l$, $l\in{\{1, \cdots, L\}}$, and define its frequency attribute as
follows.  Given a sentence of $M$ words, we use the embedding of each
word to construct a matrix, $W$, of $L$ rows and $M$ columns, whose
vertical direction records embedding values, and horizontal direction is
ordered by the word sequence. Let $w_{l,m}$ be the $(l,m)$-th element in
$W$.  A row of matrix $W$ indicates the variation of values of a
specific dimension along the sentence. By removing its mean
$\bar{w}_l=\sum_{m=1}^M w_{l,m}$, we obtain a zero-mean sequence $x_l$
where $x_{l, m}= w_{l,m}-\bar{w}_l$.  For dimension $D_l$, we use the
normalized sign-change ratio (NSR) of $x_l$ as its frequency attribute,
which can be written as
\begin{equation}
\text{NSC}(x_l) = \frac{1}{M-1} \sum_{m=1}^{M-1} \delta_{m,m+1},
\end{equation}
where $\delta_{m,m+1}=0$ if $x_{l,m}$ and $x_{l,m+1}$ are of the same sign;
otherwise, $\delta_{m,m+1}=1$. Clearly, the NSC of a dimension takes a
value between 0 and 1. Finally, we consider all sentences from a corpus,
take the average of their NSR values, and assign the averaged NSR to
each dimension as its frequency. A dimension of higher (or lower)
frequency indicates signal $x_j$ fluctuates more (or less) frequently
with respect to its mean value. 

\begin{figure}[htb]
\centerline{\includegraphics[width=6.1cm]{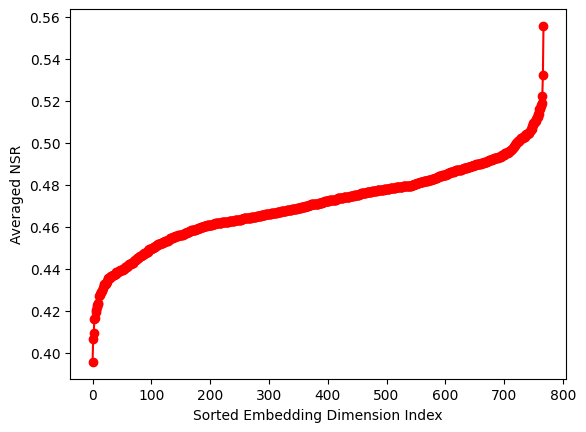}}
\caption{We plot the averaged normalized sign-change ratio (NSR) as a
function of the sorted embedding dimension index from the smallest value
($l=1$) to the largest value $l=768$) against the Penn Treebank dataset
using the BERT word embedding.  We partition dimension indices into
low-, mid-, and high-frequency sets using two elbow points with $l=50$
and $l=751$.} \label{fig:freq}
\end{figure}

We plot the averaged NSR value of sorted embedding dimension indices
against the Penn Treebank dataset using the BERT word embedding in Fig.
\ref{fig:freq}.  The dimension indices can be partitioned into low-,
mid-, and high-frequency sets using two elbow points.  They are denoted
by $S_l$, $S_m$, and $S_h$, respectively. 

\subsection{Concise Representation with Adaptive N-grams} 

We obtain the unsupervised features of a word as follows.
\begin{itemize}
\item Low-frequency dimensions \\
We examined the POS of neighboring words and observed that 92.5\% and
92.7\% of neighboring words had different POS labels in the training
sets of Penn Treebank \citep{marcus1993building} and Universal
Dependencies \citep{nivre2020universal}, respectively.  Since POS class
labels change between neighboring words in a sentence, low-frequency
embedding dimensions are not relevant to POS prediction. Thus, their
values are discarded. 
\item Mid-frequency dimensions \\
The change rates of mid-frequency dimensions are higher, making them
valuable for POS prediction and should be included in the representation
vector. The 1- and 2-grams are used for contextual and non-contextual
word embeddings, respectively, since contextual word embeddings contain
the contextual information. Additionally, we apply Principal Component
Analysis (PCA) with an energy threshold of 99\% to filter out components
corresponding to very small eigenvalues. 
\item High-frequency dimensions \\
The contextual information of a high-frequency dimension across multiple
words proves to be useful for POS prediction. This is particularly valid
for non-contextual word embedding methods (e.g., the same word ``love''
can be a verb or a noun depending on its context). It is beneficial to
use N-grams with a larger $N$ value. Since the number of high-frequency
dimensions is small, the cost is manageable. Additionally, we apply PCA
to concatenated N-gram high-frequency dimensions for dimension
reduction. 
\end{itemize}
Finally, we concatenate the N-grams from mid- and high-frequency
dimensions to get a concise representation vector of a word.

\subsection{Discriminant Feature Selection} 

The dimension of the concise representation vector of a word from the
previous step is still large.  Since not all dimensions are equally
important, it is desired to select a discriminant subset for two
purposes.  First, it can avoid the negative effects from noise or
irrelevant features.  Second, it can reduce the computational cost. For
discriminant feature selection, we adopt a supervised feature selection
method known as the discriminant feature test (DFT)
\citep{yang2022supervised}. 

DFT measures the discriminant power of each dimension of the input
vector independently. For each dimension, DFT partitions its full range
into two non-overlapping sub-intervals and uses the class labels of
training samples to compute the weighted entropy from the two
sub-intervals, called the loss function.  DFT searches over a set of
uniformly spaced points and finds the optimal point that minimizes the
loss function. Then, the minimized loss function value is assigned
to the feature as its DFT loss value. The smaller the DFT loss, the more
discriminant the associated feature. Here, we use DFT to select the most
discriminant subset of dimensions as features for POS prediction. 

\subsection{Classification for POS Labels}

After we get the discriminant features for each word, we train an
XGBoost classifier \citep{chen2016xgboost} as the target classifier
since it provides good performance and a relatively low inference
complexity as compared with other classifiers. 

\section{Experiments}\label{sec:experiments}

\subsection{Datasets and Experimental Setup} \label{sec:datasets}

{\em Datasets.} We conduct experiments on two popular English
POS tagging datasets: Penn Treebank (PTB) \citep{marcus1993building} and
Universal Dependencies (UD) \citep{nivre2020universal}.  PTB contains
material collected from the Wall Street Journal (WSJ) with 45 POS tags.
We adopt the common split of this dataset: Sections 0-18 (38,219
sentences) for training, Sections 19-21 (5,527 sentences) for
development, and Sections 22-24 (5,462 sentences) for testing.  UD
consists of 183 treebanks over 104 languages. Its English UPOS
(universal part-of-speech tags) has 17 POS tags. The default data split
is used in our experiments. 

{\em Experimental Setup.} We consider non-contextual and contextual word
embeddings with two representative examples. FastText
\citep{mikolov2018advances} is a non-contextual word embedding scheme.
The 300-dimensional FastText pre-trained on Wikipedia 2017 is used.
Fasttext utilizes subword tokenization to address the Out-of-Vocabulary
challenge, which is a serious issue in POS tagging.  BERT
\citep{kenton2019bert} is a contextual word embedding scheme. We take
the mean of embeddings of all layers as the final one. Both fastText and
BERT embeddings employ subword tokenization. In our experiments, we
utilize the mean pooling of subword embeddings as the embedding for the
associated word.  Table \ref{table:experiment_setup} lists the index
ranges of mid- and high-frequency dimensions and their N-grams.  We
choose a smaller $N$ for BERT, namely, the 1-gram for mid-frequency
dimensions and the 1-gram and 2-gram for high-frequency dimensions.
Since fastText is a non-contextual embedding, we compensate it with more
gram types.  We use DFT to choose 500 and 700 most discriminative
features for fastText and BERT embeddings, respectively.  Based on the
validation sets, the XGBoost classifier has the maximum depth equal to
3, and it has 5000 trees and 4000 trees for fastText and BERT,
respectively. 

\begin{table}[htb]
\begin{center}
\caption{Frequency partitioning and N-gram's choices}.
\label{table:experiment_setup}
\resizebox*{0.75\linewidth}{!}{
\begin{tabular}{c | c  c  c } \hline
 Word Embed. & Frequency & Indices & N-grams  \\ [0.5ex]  \hline
  \multirow{3}{*}{FastText} & Low & [0, 5] & None  \\ 
   & Mid- & [6, 260] & 1,2  \\ 
   & High- & [261, 300] & 1,2,3  \\ \hline
  \multirow{3}{*}{BERT} & Low & [0, 50] & None  \\ 
   & Mid- & [51, 750] & 1  \\ 
   & High- & [751, 768] & 1,2  \\  \hline
\end{tabular}}
\end{center}
\end{table}

\begin{table}[htb]
\caption{POS tagging accuracy on UD's test dataset.}\label{table:results_static}
\begin{center}
\resizebox*{0.55\linewidth}{!}{
\begin{tabular}{ c | c  c } \hline
 Embeddings & Fasttext & BERT \\  \hline
 MultiBPEmb & 94.30 & 96.10  \\
 GWPT (ours) & 94.94 & 96.77 \\ \hline
\end{tabular}}
\end{center}
\end{table}

\begin{table}[htb]
\caption{Comparison of model sizes and inference FLOP numbers 
of MultiBPEmb and GWPT.}\label{table:model_size}
\begin{center}
\resizebox*{1.0\linewidth}{!}{
\begin{tabular}{c | c c  c }\hline
 Methods & Modules & Param. \# & FLOPs\\ \hline
 MultiBPEmb & LSTM Layers & 3,332 K (1.55X) & 6,382 K (7.40X)\\ \hline
 \multirow{3}{*}{GWPT} & Adaptive N-gram & 281 K & 522 K \\ 
 & XGBoost & 1,870 K & 340 K \\ 
 & Total & 2,151 K (1X) &  862 K (1X) \\ \hline
\end{tabular}}
\end{center}
\end{table}

\subsection{Comparison with MultiBPEmb} \label{sec:results}

We first compare the tagging accuracy of GWPT with another word
embedding-based tagger, MultiBPEmb \citep{heinzerling2019sequence}, on
the UD dataset in Table \ref{table:results_static}.  MultiBPEmb uses two
Bi-LSTM layers and two Meta-LSTM layers with 256 hidden variables as the
POS classifier. GWPT outperforms MultiBPEmb in prediction accuracy with
both fastText and BERT embeddings. 

Next, we compare the model sizes and the computational complexity of
GWPT and MultiBPEmb in Table \ref{table:model_size}, where the inference
FLOPs (Floating-Point Operations) per word are used as the indicator of
computational complexity. Since MultiBPEmb and GWPT use the same word
embeddings (i.e., fastText or BERT), we do not include the cost of word
embeddings in the table.  Table \ref{table:model_size} shows that GWPT
has smaller model size and lower inference computational complexity than
MultiBPEmb. The estimated model size and inference FLOPs for GWPT using
fastText embedding on the UD dataset is given below. The main components
that contribute to the model size and computational complexity in GWPT's
inference are adaptive N-grams and XGBoost. Other components, say,
frequency partitioning and discriminant feature Selection, have
negligible parameter counts and computational complexity. 

\begin{figure}[htb]
\begin{center}
\includegraphics[width=0.7\linewidth]{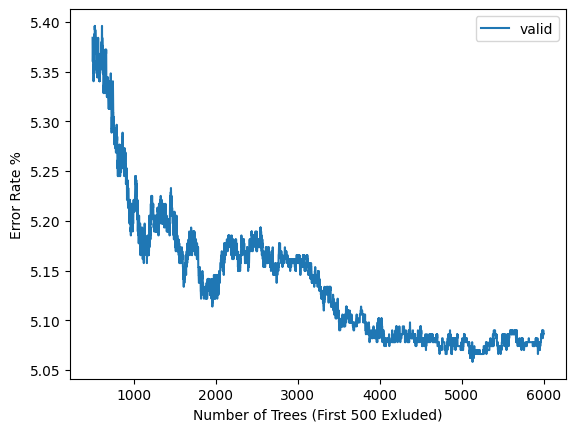} \\
\includegraphics[width=0.7\linewidth]{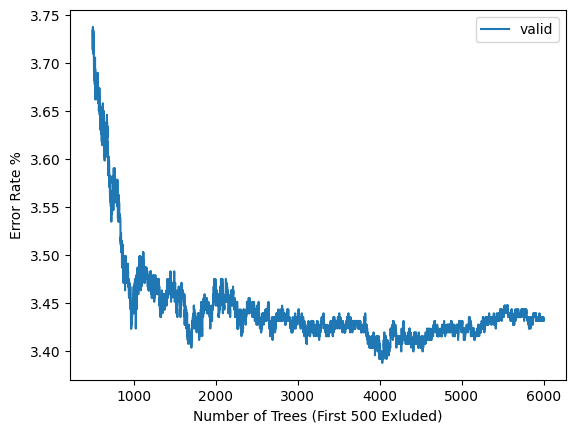}
\end{center}
\caption{The validation error rate as a function of the XGBoost tree
numbers for each class on the UD datasets: (top) fastText and (bottom)
BERT.}\label{fig:xgb}
\end{figure}

\begin{itemize}
\item Adaptive N-grams. PCA is applied to N-grams. Since the
mid-frequency range (from indices 5 to 260) encompasses 255 dimensions
and involves 2-gram features, the parameter count for PCA is less than
$(255 \times 2)^2 = 260,100$. The high-frequency range (from indices 261
to 300) contains 40 dimensions and involves both 2-grams and 3-grams,
and the parameter number for PCA is less than $(40 \times 2)^2 + (40
\times 3)^2 = 20,800$. Thus, the total is bounded by $280,900$.  The
FLOPs for a PCA transform is $ 2 \times m \times n $ where $m$ and $n$
are input and output dimensions, respectively. Thus, the FLOPs is $ 2
\times (490 \times 490 + 80 \times 80 + 120 \times 120) = 521,800$

\item XGBoost. A tree with a depth of 3 has 22 parameters. In multiclass
classification problems, XGBoost employs the One versus Rest strategy.
We use validation sets to select the tree number for each class in
XGBoost. Fig. \ref{fig:xgb} illustrates the relationship between the
validation error rate and the number of trees for each class in XGBoost.
The error rates of the first 500 trees are excluded for better
visualization.  the validation error rates converge at 5,000 and 4,000
trees in each class for fastText and BERT embeddings, respectively. The
UD dataset has 17 classes of POS. Thus, the total number of parameters
for fastText is $5,000 \times 22 \times 17 = 1,870,000 $. The FLOPs for
an XGBoost classifier in each class are the number of trees times the
tree depth. All trees' predictions need to be summed up via addition.
Thus, the FLOPs is $(5,000 \times 3 + 5,000) \times 17 = 340,000$. 

\end{itemize}

\subsection{Comparison with Other POS Taggers} \label{sec:results2}

We further compare the performance with other POS taggers for PTB and UD
in Table \ref{table:results_contextual}. Meta-BiLSTM
\citep{bohnet-etal-2018-morphosyntactic}, Char Bi-LSTM
\citep{ling2015finding} and Adversarial Bi-LSTM
\citep{yasunaga2018robust} are LSTM models built on character and
word-based representations.  BiLSTM-LAN \citep{cui2019hierarchically} is
a multi-layered BiLSTM-softmax sequence labeler with an attention
mechanism.  Flair embeddings \citep{akbik2018contextual} adopts the
character embedding. In addition, we fine-tune the whole BERT model with
extra linear layers for POS tagging, which is denoted as BERT-MLP.  We
see that our method can still achieve competitive performance without
character-level information and complicated training strategies. 

\begin{table}[htb]
\caption{Comparison of POS tagging accuracy rate for PTB and UD test
datasets, where [\textsuperscript{\textdagger}] denotes a method
implemented by ourselves.}\label{table:results_contextual}
\begin{center}
\resizebox*{0.65\linewidth}{!}{
\begin{tabular}{ c | c | c} \hline
 Methods & PTB & UD \\ \hline
  Meta BiLSTM & \textbf{97.96} & - \\ 
  Flair embeddings & 97.85 & - \\
  Char Bi-LSTM & 97.78 & - \\
  BiLSTM-LAN & 97.65 & 95.59\\
  Adversarial Bi-LSTM & 97.58 & 95.82 \\
  BERT-MLP \textsuperscript{\textdagger} & 97.67 & 96.32 \\  \hline
  GWPT/BERT (Ours) & 97.73 & \textbf{96.77} \\  \hline
\end{tabular}}
\end{center}
\end{table}

\begin{table}[htb]
\caption{POS tagging accuracy using different N-grams for the UD 
dataset.}\label{table:ablation_freq}
\begin{center}
\resizebox*{0.80\linewidth}{!}{
\begin{tabular}{c | c  c  c }  \hline
 Word Embed. & N-grams  & Feature Dim. & Accuracy \\ \hline
  \multirow{4}{*}{FastText} & 1 & 300 & 88.56 \\ 
    & 1, 2 & 1.5K & 94.52 \\ 
    & 1, 2, 3 & 4K & 94.82 \\ 
    & adaptive & 2K & 94.80 \\ \hline
  \multirow{4}{*}{BERT} & 1 & 0.7 k & 96.64 \\ 
   & 1,2 & 3.5K & 96.72 \\ 
   & 1,2,3 & 9.6K & 96.64 \\ 
   & adaptive & 0.7K & 96.72 \\  \hline
\end{tabular}}
\end{center}
\end{table}

\subsection{Ablation Study}

We conduct ablation studies to illustrate the effects of adaptive
N-grams and DFT. 

\textbf{Adaptive N-grams.} We compare the performance of two settings:
1) fixed N-grams for all dimensions of word embeddings, and 2) the
proposed adaptive N-grams.  in Table \ref{table:ablation_freq}.
FastText achieves its best performance using up to 3-grams. BERT
embeddings require only 2-grams to boost the performance due to their
inherent contextual information.  Increasing the neighboring context,
such as 3-grams, conversely impacts the results. Our adaptive n-grams
achieves similar performance but with significantly reduced feature
dimensions. 

\begin{figure}[htb]
\centerline{\includegraphics[width=0.85\linewidth]{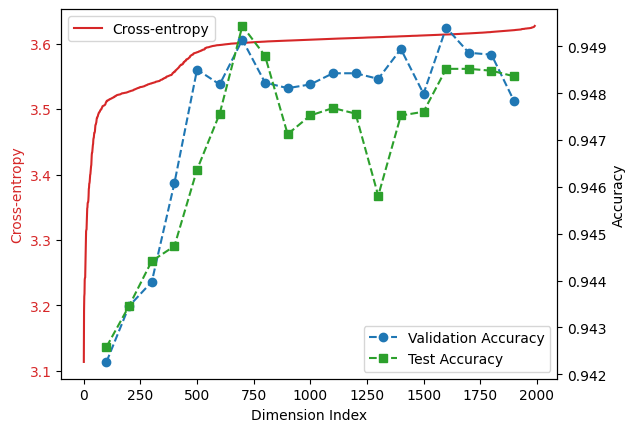}}
\caption{Sorted discriminability for each feature dimension selected by DFT and
validation and test accuracies on the UD dataset. A lower cross-entropy value indicates
a more discriminant feature.} \label{fig:dft}
\end{figure}

\textbf{DFT.} 
Fig. \ref{fig:dft} shows the curves of sorted discriminability (i.e.,
cross-entropy) for each feature dimension of word representation derived
from fastText for the UD dataset. Within the same figure, we depict the
validation and test accuracies for POS tagging using all the features
selected by DFT up to the dimension index in the x-axis. We see
consistent classification performance with the feature discriminability. 
Furthermore, we compare the performance of using the original adaptive n-gram
features and the discriminative features selected by DFT in Table
\ref{table:ablation_dft}. It shows that the POS tagging accuracy can be
further improved by removing irrelevant or noisy features using DFT. 

\begin{table}[htb]
\caption{POS tagging accuracy using DFT on the UD test set}\label{table:ablation_dft}
\begin{center}
\resizebox*{0.80\linewidth}{!}{
\begin{tabular}{ c | c  c  c }
 \hline
 Word Embed. & Features & Dimension & Accuracy  \\ \hline
  \multirow{2}{*}{FastText} & Before DFT & 1992 & 94.80 \\ 
   & After DFT & 500 & 94.94 \\ \hline
   \multirow{2}{*}{BERT} & Before DFT & 733 & 96.72  \\ 
    & After DFT & 700 & 96.77\\ \hline
\end{tabular}}
\end{center}
\end{table}

\subsection{Effect of Parameters in XGBoost}

\begin{figure}[htb]
\centerline{\includegraphics[width=0.86\linewidth]{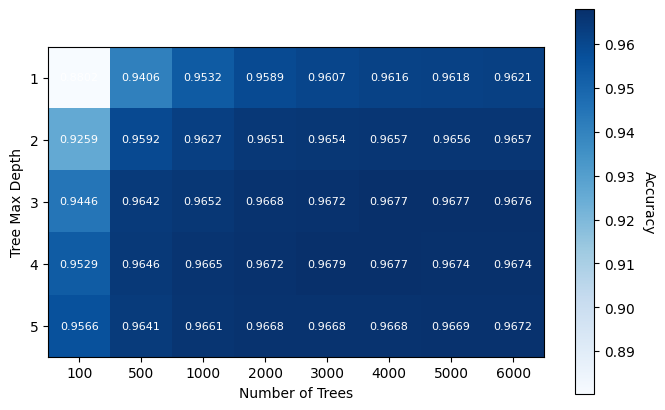}}
\centerline{\includegraphics[width=0.86\linewidth]{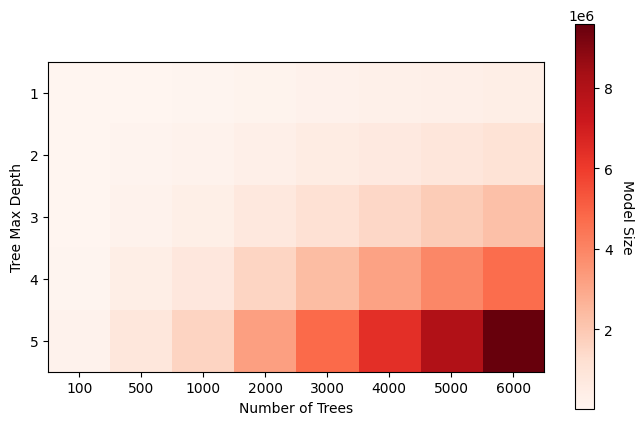}}
\caption{The effect of the maximum depth and the tree number in XGBoost
on GWPT for the UD test set: POS tagging accuracy (top) and the model
size (bottom).} \label{fig:xgb_params}
\end{figure}

We studied the impact of two important parameters for the XGBoost
classifier: the maximum depth and the tree number. Figure
\ref{fig:xgb_params} illustrates the performance of GWPT on the UD test
dataset (top) and the model size of different tree maximum depths and
tree numbers (bottom). Although GWPT's performance improves as the tree
maximum depth and the tree number increase.  The model size grows
greatly once the tree maximum depth is larger than 2 and the tree number
is greater than 2000 while the improvement in accuracy is marginal.  For
this reason, we carefully set the maximum depth to 3 and the tree number
to 4000 in order to strike a balance between performance and model
size/complexity. 

\section{Conclusion and Future Work}\label{sec:conclu}

A novel lightweight word-embedding-based POS Tagger, called GWPT, was
proposed in this work. GWPT was designed with a modular structure. It
analyzed word embedding frequencies, employed adaptive N-grams based on
frequency intervals, selected discriminative features, and adopted the
XGBoost classifier. It offered competitive POS tagging performance with
few parameters and much lower inference complexity. 

As future extensions, we can exploit character embedding to boost the
performance further. Additionally, the XGBoost classifier is not
effective in handling multi-class classification problems since its
model sizes increase rapidly. It would be interesting to design more
efficient and lightweight classifiers for GWPT. 

\bibliographystyle{plainnat}
\bibliography{mybibfile}

\begin{thebibliography}{38}
\providecommand{\natexlab}[1]{#1}
\providecommand{\url}[1]{\texttt{#1}}
\expandafter\ifx\csname urlstyle\endcsname\relax
  \providecommand{\doi}[1]{doi: #1}\else
  \providecommand{\doi}{doi: \begingroup \urlstyle{rm}\Url}\fi

\bibitem[Agarwal and Mani(2006)]{agarwal2006part}
Himanshu Agarwal and Anirudh Mani.
\newblock Part of speech tagging and chunking with conditional random fields.
\newblock In \emph{the Proceedings of NWAI workshop}, 2006.

\bibitem[Akbik et~al.(2018)Akbik, Blythe, and Vollgraf]{akbik2018contextual}
Alan Akbik, Duncan Blythe, and Roland Vollgraf.
\newblock Contextual string embeddings for sequence labeling.
\newblock In \emph{Proceedings of the 27th international conference on computational linguistics}, pages 1638--1649, 2018.

\bibitem[Bohnet et~al.(2018{\natexlab{a}})Bohnet, McDonald, Sim{\~o}es, Andor, Pitler, and Maynez]{bohnet-etal-2018-morphosyntactic}
Bernd Bohnet, Ryan McDonald, Gon{\c{c}}alo Sim{\~o}es, Daniel Andor, Emily Pitler, and Joshua Maynez.
\newblock Morphosyntactic tagging with a meta-{B}i{LSTM} model over context sensitive token encodings.
\newblock In \emph{Proceedings of the 56th Annual Meeting of the Association for Computational Linguistics (Volume 1: Long Papers)}, pages 2642--2652, Melbourne, Australia, July 2018{\natexlab{a}}. Association for Computational Linguistics.
\newblock \doi{10.18653/v1/P18-1246}.
\newblock URL \url{https://aclanthology.org/P18-1246}.

\bibitem[Bohnet et~al.(2018{\natexlab{b}})Bohnet, McDonald, Sim{\~o}es, Andor, Pitler, and Maynez]{bohnet2018morphosyntactic}
Bernd Bohnet, Ryan McDonald, Gon{\c{c}}alo Sim{\~o}es, Daniel Andor, Emily Pitler, and Joshua Maynez.
\newblock Morphosyntactic tagging with a meta-bilstm model over context sensitive token encodings.
\newblock In \emph{Proceedings of the 56th Annual Meeting of the Association for Computational Linguistics (Volume 1: Long Papers)}, pages 2642--2652, 2018{\natexlab{b}}.

\bibitem[Brill(1992)]{brill1992simple}
Eric Brill.
\newblock A simple rule-based part of speech tagger.
\newblock In \emph{Speech and Natural Language: Proceedings of a Workshop Held at Harriman, New York, February 23-26, 1992}, 1992.

\bibitem[Brown et~al.(2020)Brown, Mann, Ryder, Subbiah, Kaplan, Dhariwal, Neelakantan, Shyam, Sastry, Askell, et~al.]{brown2020language}
Tom Brown, Benjamin Mann, Nick Ryder, Melanie Subbiah, Jared~D Kaplan, Prafulla Dhariwal, Arvind Neelakantan, Pranav Shyam, Girish Sastry, Amanda Askell, et~al.
\newblock Language models are few-shot learners.
\newblock \emph{Advances in neural information processing systems}, 33:\penalty0 1877--1901, 2020.

\bibitem[Chen et~al.(2021)Chen, Rouhsedaghat, Ghani, Hu, You, and Kuo]{chen2021defakehop}
Hong-Shuo Chen, Mozhdeh Rouhsedaghat, Hamza Ghani, Shuowen Hu, Suya You, and C-C~Jay Kuo.
\newblock Defakehop: A light-weight high-performance deepfake detector.
\newblock In \emph{2021 IEEE International conference on Multimedia and Expo (ICME)}, pages 1--6. IEEE, 2021.

\bibitem[Chen and Guestrin(2016)]{chen2016xgboost}
Tianqi Chen and Carlos Guestrin.
\newblock Xgboost: A scalable tree boosting system.
\newblock In \emph{Proceedings of the 22nd acm sigkdd international conference on knowledge discovery and data mining}, pages 785--794, 2016.

\bibitem[Chen et~al.(2020)Chen, Rouhsedaghat, You, Rao, and Kuo]{chen2020pixelhop++}
Yueru Chen, Mozhdeh Rouhsedaghat, Suya You, Raghuveer Rao, and C-C~Jay Kuo.
\newblock Pixelhop++: A small successive-subspace-learning-based (ssl-based) model for image classification.
\newblock In \emph{2020 IEEE International Conference on Image Processing (ICIP)}, pages 3294--3298. IEEE, 2020.

\bibitem[Chiche and Yitagesu(2022)]{chiche2022part}
Alebachew Chiche and Betselot Yitagesu.
\newblock Part of speech tagging: a systematic review of deep learning and machine learning approaches.
\newblock \emph{Journal of Big Data}, 9\penalty0 (1):\penalty0 1--25, 2022.

\bibitem[Cui and Zhang(2019)]{cui2019hierarchically}
Leyang Cui and Yue Zhang.
\newblock Hierarchically-refined label attention network for sequence labeling.
\newblock In \emph{Proceedings of the 2019 Conference on Empirical Methods in Natural Language Processing and the 9th International Joint Conference on Natural Language Processing (EMNLP-IJCNLP)}, pages 4115--4128, 2019.

\bibitem[Eric(1994)]{eric1994some}
Brill Eric.
\newblock Some advances in transformation-based part of speech tagging.
\newblock \emph{Proceedings of the Twelveth AAAI, 1994}, 1994.

\bibitem[Heinzerling and Strube(2019)]{heinzerling2019sequence}
Benjamin Heinzerling and Michael Strube.
\newblock Sequence tagging with contextual and non-contextual subword representations: A multilingual evaluation.
\newblock In \emph{Proceedings of the 57th Annual Meeting of the Association for Computational Linguistics}, pages 273--291, 2019.

\bibitem[Kadam et~al.(2022)Kadam, Zhang, Liu, and Kuo]{kadam2022r}
Pranav Kadam, Min Zhang, Shan Liu, and C-C~Jay Kuo.
\newblock R-pointhop: A green, accurate, and unsupervised point cloud registration method.
\newblock \emph{IEEE Transactions on Image Processing}, 31:\penalty0 2710--2725, 2022.

\bibitem[Kenton and Toutanova(2019)]{kenton2019bert}
Jacob Devlin Ming-Wei~Chang Kenton and Lee~Kristina Toutanova.
\newblock Bert: Pre-training of deep bidirectional transformers for language understanding.
\newblock In \emph{Proceedings of NAACL-HLT}, pages 4171--4186, 2019.

\bibitem[Kim et~al.(1999)Kim, Lee, and Rim]{kim1999hmm}
Jin-Dong Kim, Sang-Zoo Lee, and Hae~Chang Rim.
\newblock Hmm specialization with selective lexicalization.
\newblock In \emph{1999 Joint SIGDAT Conference on Empirical Methods in Natural Language Processing and Very Large Corpora}, 1999.

\bibitem[Kuo(2016)]{kuo2016understanding}
C-C~Jay Kuo.
\newblock Understanding convolutional neural networks with a mathematical model.
\newblock \emph{Journal of Visual Communication and Image Representation}, 41:\penalty0 406--413, 2016.

\bibitem[Kuo and Chen(2018)]{kuo2018data}
C-C~Jay Kuo and Yueru Chen.
\newblock On data-driven saak transform.
\newblock \emph{Journal of Visual Communication and Image Representation}, 50:\penalty0 237--246, 2018.

\bibitem[Kuo and Madni(2022)]{kuo2022green}
C-C~Jay Kuo and Azad~M Madni.
\newblock Green learning: Introduction, examples and outlook.
\newblock \emph{Journal of Visual Communication and Image Representation}, page 103685, 2022.

\bibitem[Kuo et~al.(2019)Kuo, Zhang, Li, Duan, and Chen]{kuo2019interpretable}
C-C~Jay Kuo, Min Zhang, Siyang Li, Jiali Duan, and Yueru Chen.
\newblock Interpretable convolutional neural networks via feedforward design.
\newblock \emph{Journal of Visual Communication and Image Representation}, 60:\penalty0 346--359, 2019.

\bibitem[Lee et~al.(2000)Lee, Tsujii, and Rim]{lee2000lexicalized}
Sang-Zoo Lee, Jun’ichi Tsujii, and Hae~Chang Rim.
\newblock Lexicalized hidden markov models for part-of-speech tagging.
\newblock In \emph{COLING 2000 Volume 1: The 18th International Conference on Computational Linguistics}, 2000.

\bibitem[Li et~al.(2021)Li, Mao, and Wang]{li2021part}
Hongwei Li, Hongyan Mao, and Jingzi Wang.
\newblock Part-of-speech tagging with rule-based data preprocessing and transformer.
\newblock \emph{Electronics}, 11\penalty0 (1):\penalty0 56, 2021.

\bibitem[Ling et~al.(2015)Ling, Dyer, Black, Trancoso, Fermandez, Amir, Marujo, and Lu{\'\i}s]{ling2015finding}
Wang Ling, Chris Dyer, Alan~W Black, Isabel Trancoso, Ram{\'o}n Fermandez, Silvio Amir, Luis Marujo, and Tiago Lu{\'\i}s.
\newblock Finding function in form: Compositional character models for open vocabulary word representation.
\newblock In \emph{Proceedings of the 2015 Conference on Empirical Methods in Natural Language Processing}, pages 1520--1530, 2015.

\bibitem[Marcus et~al.(1993)Marcus, Santorini, and Marcinkiewicz]{marcus1993building}
Mitchell Marcus, Beatrice Santorini, and Mary~Ann Marcinkiewicz.
\newblock Building a large annotated corpus of english: The penn treebank.
\newblock \emph{University of Pennsylvania, Department of Computer and Information Science Technical Report No. MS-CIS-93-87}, 1993.

\bibitem[Mikolov et~al.(2018)Mikolov, Grave, Bojanowski, Puhrsch, and Joulin]{mikolov2018advances}
Tomas Mikolov, Edouard Grave, Piotr Bojanowski, Christian Puhrsch, and Armand Joulin.
\newblock Advances in pre-training distributed word representations.
\newblock In \emph{Proceedings of the International Conference on Language Resources and Evaluation (LREC 2018)}, 2018.

\bibitem[Nivre et~al.(2020)Nivre, de~Marneffe, Ginter, Hajic, Manning, Pyysalo, Schuster, Tyers, and Zeman]{nivre2020universal}
Joakim Nivre, Marie-Catherine de~Marneffe, Filip Ginter, Jan Hajic, Christopher~D Manning, Sampo Pyysalo, Sebastian Schuster, Francis Tyers, and Daniel Zeman.
\newblock Universal dependencies v2: An evergrowing multilingual treebank collection.
\newblock In \emph{Proceedings of the Twelfth Language Resources and Evaluation Conference}, pages 4034--4043, 2020.

\bibitem[OpenAI(2023)]{openai2023gpt}
R~OpenAI.
\newblock Gpt-4 technical report.
\newblock \emph{arXiv}, pages 2303--08774, 2023.

\bibitem[PVS and Karthik(2007)]{pvs2007part}
Avinesh PVS and G~Karthik.
\newblock Part-of-speech tagging and chunking using conditional random fields and transformation based learning.
\newblock \emph{Shallow parsing for south asian languages}, 21\penalty0 (21-24):\penalty0 2, 2007.

\bibitem[Ratnaparkhi(1996)]{ratnaparkhi1996maximum}
Adwait Ratnaparkhi.
\newblock A maximum entropy model for part-of-speech tagging.
\newblock In \emph{Conference on empirical methods in natural language processing}, 1996.

\bibitem[Silfverberg et~al.(2014)Silfverberg, Ruokolainen, Lind{\'e}n, and Kurimo]{silfverberg2014part}
Miikka Silfverberg, Teemu Ruokolainen, Krister Lind{\'e}n, and Mikko Kurimo.
\newblock Part-of-speech tagging using conditional random fields: Exploiting sub-label dependencies for improved accuracy.
\newblock In \emph{Proceedings of the 52nd Annual Meeting of the Association for Computational Linguistics (Volume 2: Short Papers)}, pages 259--264, 2014.

\bibitem[Van~Gael et~al.(2009)Van~Gael, Vlachos, and Ghahramani]{van2009infinite}
Jurgen Van~Gael, Andreas Vlachos, and Zoubin Ghahramani.
\newblock The infinite hmm for unsupervised pos tagging.
\newblock In \emph{Proceedings of the 2009 Conference on Empirical Methods in Natural Language Processing}, pages 678--687, 2009.

\bibitem[Wang et~al.(2015{\natexlab{a}})Wang, Qian, Soong, He, and Zhao]{wang2015part}
Peilu Wang, Yao Qian, Frank~K Soong, Lei He, and Hai Zhao.
\newblock Part-of-speech tagging with bidirectional long short-term memory recurrent neural network.
\newblock \emph{arXiv preprint arXiv:1510.06168}, 2015{\natexlab{a}}.

\bibitem[Wang et~al.(2015{\natexlab{b}})Wang, Qian, Soong, He, and Zhao]{wang2015unified}
Peilu Wang, Yao Qian, Frank~K Soong, Lei He, and Hai Zhao.
\newblock A unified tagging solution: Bidirectional lstm recurrent neural network with word embedding.
\newblock \emph{arXiv preprint arXiv:1511.00215}, 2015{\natexlab{b}}.

\bibitem[Wei et~al.(2023)Wei, Wang, Wang, and Kuo]{wei2023overview}
Chengwei Wei, Yun-Cheng Wang, Bin Wang, and C-C~Jay Kuo.
\newblock An overview on language models: Recent developments and outlook.
\newblock \emph{arXiv preprint arXiv:2303.05759}, 2023.

\bibitem[Yang et~al.(2022)Yang, Wang, Fu, Kuo, et~al.]{yang2022supervised}
Yijing Yang, Wei Wang, Hongyu Fu, C-C~Jay Kuo, et~al.
\newblock On supervised feature selection from high dimensional feature spaces.
\newblock \emph{APSIPA Transactions on Signal and Information Processing}, 11\penalty0 (1), 2022.

\bibitem[Yasunaga et~al.(2018)Yasunaga, Kasai, and Radev]{yasunaga2018robust}
Michihiro Yasunaga, Jungo Kasai, and Dragomir Radev.
\newblock Robust multilingual part-of-speech tagging via adversarial training.
\newblock In \emph{Proceedings of NAACL-HLT}, pages 976--986, 2018.

\bibitem[Zhang et~al.(2020)Zhang, You, Kadam, Liu, and Kuo]{zhang2020pointhop}
Min Zhang, Haoxuan You, Pranav Kadam, Shan Liu, and C-C~Jay Kuo.
\newblock Pointhop: An explainable machine learning method for point cloud classification.
\newblock \emph{IEEE Transactions on Multimedia}, 22\penalty0 (7):\penalty0 1744--1755, 2020.

\bibitem[Zhao and Wang(2002)]{zhao2002chinese}
Jian Zhao and Xiao-long Wang.
\newblock Chinese pos tagging based on maximum entropy model.
\newblock In \emph{Proceedings. International Conference on Machine Learning and Cybernetics}, volume~2, pages 601--605. IEEE, 2002.

\end{thebibliography}

\end{document}